\documentclass{article} 
\usepackage{iclr2026_conference,times}

\usepackage{amsmath,amsfonts,bm}

\def\eqref#1{equation~\ref{#1}}

\def\1{\bm{1}}

\DeclareMathAlphabet{\mathsfit}{\encodingdefault}{\sfdefault}{m}{sl}
\SetMathAlphabet{\mathsfit}{bold}{\encodingdefault}{\sfdefault}{bx}{n}

\usepackage{hyperref}
\usepackage{url}
\usepackage{times}
\usepackage{helvet}
\usepackage{courier}
\usepackage{graphicx}
\usepackage{algorithm}
\usepackage{amsmath}
\usepackage{algpseudocode}
\usepackage{caption}
\usepackage{algorithmicx}
\usepackage{arydshln}
\usepackage{amsfonts}
\usepackage{xcolor}
\usepackage{newfloat}
\usepackage{listings}
\usepackage{subcaption}
\usepackage{wrapfig}
\usepackage{amsthm}  
\usepackage{amsthm}
\usepackage{amssymb}
\theoremstyle{plain}
\newtheorem{theorem}{Theorem}
\newtheorem{lemma}{Lemma}

\theoremstyle{remark}

\usepackage[most]{tcolorbox}
\tcbset{
  remarkstyle/.style={
    colback=gray!3,
    colframe=gray!40,
    boxrule=0.4pt,
    arc=2pt,
    left=6pt,
    right=6pt,
    top=4pt,
    bottom=4pt,
    before skip=6pt,
    after skip=6pt
  }
}
\newtcolorbox{remarkbox}[1][]{remarkstyle,title=\textbf{Remark},#1}

\floatstyle{ruled}
\newfloat{listing}{tb}{lst}{}
\floatname{listing}{Listing}

\iclrfinalcopy 

\title{ PIXEL: Adaptive Steering Via Position-wise Injection with eXact Estimated Levels under Subspace Constraint}

\author{
\begin{tabular}{l}
Manjiang Yu$^{1,*,\dagger}$,
Hongji Li$^{2,3,4*}$,
Priyanka Singh$^{1}$,
Xue Li$^{1}$,
Di Wang$^{3,4}$,
Lijie Hu$^{2,\dagger}$
\end{tabular}
\\[0.3em]
$^1$ University of Queensland, Brisbane, Australia \\
$^2$ Mohamed bin Zayed University of Artificial Intelligence (MBZUAI) \\
$^3$ Provable Responsible AI and Data Analytics (PRADA) Lab \\
$^4$ King Abdullah University of Science and Technology (KAUST) \\
}

\begin{document}

\title{PIXEL: Adaptive Steering Via Position-wise Injection with eXact Estimated Levels under Subspace Calibration}
\maketitle


\begin{abstract}
Reliable behavior control is central to deploying Large Language Models (LLMs) on the web. Activation steering offers a tuning-free route to align attributes (e.g., truthfulness) that ensure trustworthy generation. Prevailing approaches rely on coarse heuristics and lack a principled account of where to steer and how strongly to intervene. To this end, we propose \underline{P}osition-wise \underline{I}njection with e\underline{X}act \underline{E}stimated \underline{L}evels (PIXEL), a position-wise activation steering framework that, in contrast to prior work, learns a property-aligned subspace from dual views (tail-averaged and end-token) and selects intervention strength via a constrained geometric objective with a closed-form solution, thereby adapting to token-level sensitivity without global hyperparameter tuning. PIXEL further performs sample-level orthogonal residual calibration to refine the global attribute direction and employs a lightweight position-scanning routine to identify receptive injection sites. We additionally provide representation-level guarantees for the minimal-intervention rule, supporting reliable alignment. Across diverse models and evaluation paradigms, PIXEL consistently improves attribute alignment while preserving model general capabilities, offering a practical and principled method for LLMs' controllable generation. Our code is available at \url{https://github.com/V1centNevwake/PIXEL-Adaptive-Steering}

\end{abstract}

\def\thefootnote{*}\footnotetext{Equal Contribution.}
\def\thefootnote{†}\footnotetext{Corresponding Author.}


\section{Introduction}

Large Language Models (LLMs)~\citep{brown2020language, touvron2023llama, openai2023gpt} have achieved remarkable success across a broad spectrum of natural language processing tasks, ranging from question answering to open-ended generation~\cite{su2023detectllm,su2023fake,cheng2024multi,yang2024model,zhang2024locate,yang2025understanding,hu2024differentially}. 
However, ensuring that LLMs behave in alignment with desired attributes such as truthfulness, fairness, or informativeness remains an open challenge, especially in real-world deployments where outputs must satisfy domain-specific or ethical constraints~\citep{ouyang2022training, bai2022constitutional,yang2024dialectical,zhang2025understanding,wang2025truth,zhou2025flattery,dong2025understanding,yang2025fraud,guo2025benchmarking}. 

To meet this need, a growing body of work has focused on post-training control mechanisms, which aim to adjust model behavior without retraining the entire model. 
Among them, activation steering has emerged as a promising lightweight approach that manipulates internal representations at inference time to guide outputs toward target properties~\citep{rimsky2024steering, li2023steering,yang2024exploring,hu2024hopfieldian}. 
Compared to supervised fine-tuning~\citep{wei2022finetuned}(SFT) or reinforcement learning with human feedback~\citep{christiano2017deep}(RLHF), steering methods require no additional gradient updates and can be flexibly applied to frozen LLM. 
Most existing methods construct steering based on the activation difference between positive and negative samples and apply a shift with a strength controlled by a coefficient at selected layers or locations.

While effective, these methods are typically limited by two core challenges. First, these methods typically apply a fixed-size steering vector to each location, ignoring the fact that different layers and tokens are responsive to interventions to varying degrees. In fact, recent work CAE~\citep{hao2025patterns_of_cae} shows that steering vector injection can harm model perplexity if it is too broad or too strong. Second, interventions are often applied indiscriminately across all layers or based on hand-picked locations without a principled understanding of where steering is most effective. This coarse-grained design limits the flexibility and reliability of activation-based steering and can degrade the power of general models when misapplied~\citep{sheng2025alphasteer}.

Recent work has introduced adaptability into activation control. ACT ~\citep{wang2024adaptive}trains probes and clusters direction vectors into categories, then scales the edits based on the probe score of the selected head, adjusting the strength of the intervention at inference. SADI ~\citep{zhang2025sadi} dynamically adjusts the control direction by performing element-wise edits on the connection activations of all layers, further adjusting the intervention based on the input semantics. However, neither approach provides a principled treatment of the location and extent of the intervention. ACT relies on learned detectors and task-tuned hyperparameters, which limit its robustness and transferability across attribute-aligned datasets. While SADI does not share ACT's limitations, it requires uniform application of interventions across all layers, preventing adaptive selection of the location of injected steering vectors.

To overcome these limitations, we introduce the \textbf{\underline{P}osition-wise \underline{I}njection with e\underline{X}act \underline{E}stimated \underline{L}evels (PIXEL)}, a framework that performs fine-grained, location-wise adaptive control with minimal intervention. Specifically, we combine the tail-averaged view with the end-token view to learn a two-view, attribute-aligned subspace, providing a robust steering foundation based on validated samples. We then transform steering strength selection into a constrained geometric optimization and derive a closed-form solution that yields the minimum edits required to achieve target alignment at each labeled position, thereby generating a positional steering blueprint that requires no global hyperparameter tuning. We further incorporate orthogonal residual calibration to adapt to sample-specific semantics while preserving the global attribute subspace, and we employ a lightweight dynamic positional scanning routine to select recipient injection points (e.g., single-layer or continuous multi-layer injection). Furthermore, our representation-level analysis provides theoretically supported guarantees, including the non-negativity of site-alignment margins with minimal per-site intervention and generalization bounds on the normalized mean margin.

We validate our approach on a variety of backbone networks and evaluation formats, including Llama3-8B-Instruct, Qwen2-7B-Instruct, and Mistral-7B-v0.3, as well as on benchmarks covering multiple-choice(TruthfulQA, BBQ) and open-ended generation(Refusal and HelpSteer). Our experimental results reveal that PIXEL significantly outperforms existing activation intervention methods.

Our contributions can be summarized as follows:
\begin{itemize}

\item We construct a dual-view, attribute-aligned subspace by jointly leveraging tail-averaged and end-token representations and derive a closed-form, per-position minimum steering strength, eliminating empirical adjustments while ensuring principled control over the magnitude of intervention.
\item We introduce a position-guided blueprint, combined with a lightweight dynamic position scanning routine, that selectively injects at the most receptive layers and labels to maximize control efficiency while minimizing unnecessary intervention.
\item We demonstrate significant improvements across different model backbones and evaluation paradigms, including multiple-choice settings and open-ended generation. We also test that after steering, the model maintains state-of-the-art performance on common performance benchmarks (e.g., RACE, MMLU).

\end{itemize}

\begin{figure*}[t]
\centering
\includegraphics[width=\textwidth, trim={10 50 10 50}, clip]{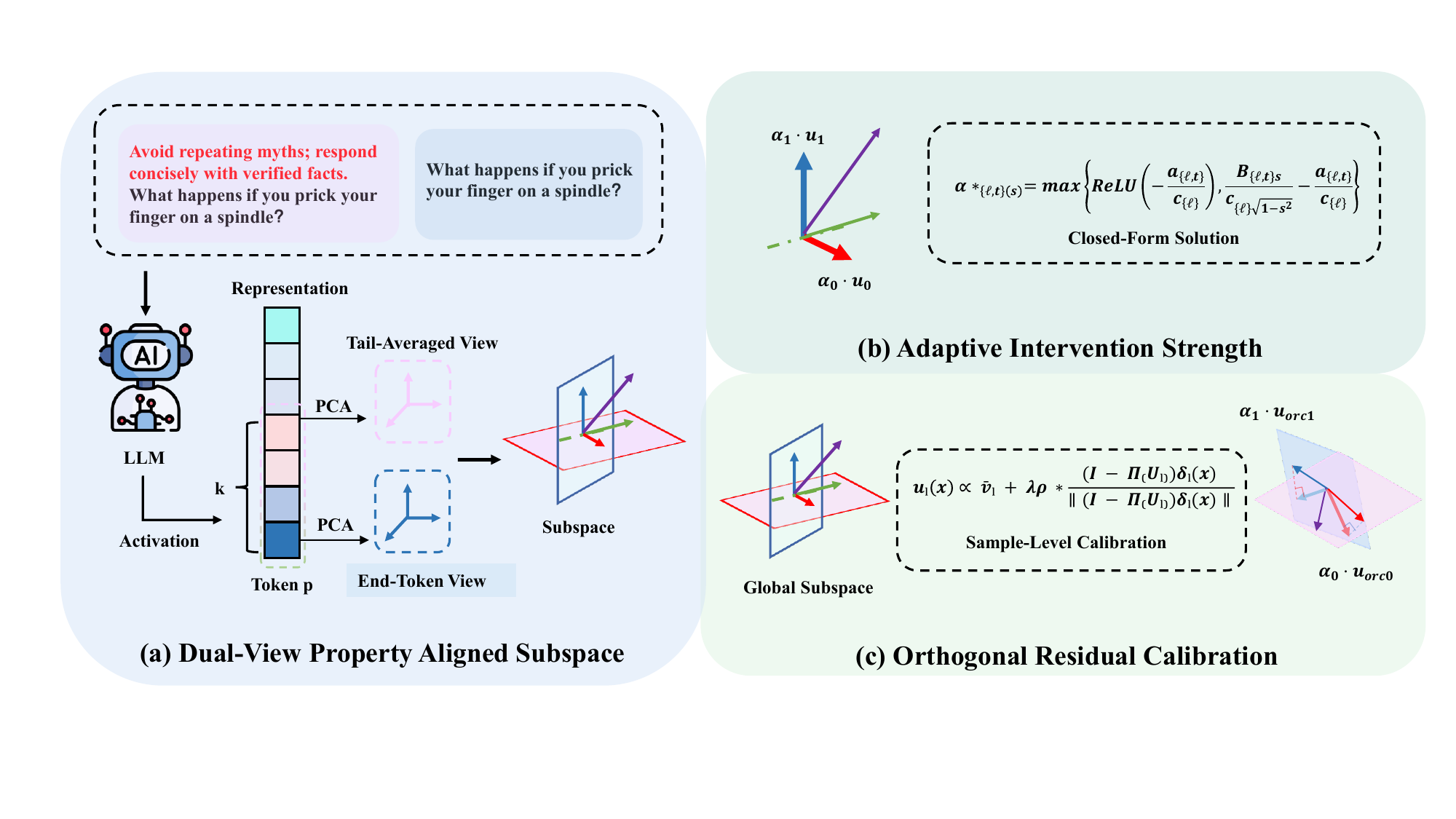}
\caption{
Overview of \textsc{PIXEL}. 
(a) \textbf{Dual-View Property-Aligned Subspace}: Tail-averaged and end-token differentials from validated, property-aligned samples are projected via PCA to form a robust attribute-aligned subspace (Sec.~\ref{sec:subspace}). 
(b) \textbf{Adaptive Intervention Strength}: A closed-form, per-position minimum steering strength is derived to avoid global hyperparameter tuning (Sec.~\ref{sec:adaptive_alpha}).
(c) \textbf{Orthogonal Residual Calibration}: The global attribute direction is refined by a sample-specific orthogonal residual, enabling context-aware alignment while maintaining global consistency (Sec.~\ref{sec:adaptive_alpha}).
}
\label{fig:overview}
\end{figure*}

\section{Related Works}
\label{related work}

\subsection{Activation Steering Methods} 
Activation steering refers to modifying hidden representations of large language models at inference time to adjust outputs toward desired attributes, without updating model parameters~\citep{im2025unified,NEURIPS2024_58cbe393,bayat2025steering,oozeer2025beyond,subramani2024steering,liu2024steerability,hernandez2023linearity,yang2025d,yao2025understanding,jiang2025msrs}. 
A variety of methods have been proposed in this line of work. 
Contrastive Activation Addition (CAA)~\citep{rimsky-etal-2024-steering} computes steering vectors by averaging activation differences between positive and negative examples and injects them at inference to shift model behavior. 
Inference-Time Intervention (ITI)~\citep{NEURIPS2023_81b83900} applies fixed steering directions across attention heads to enhance factuality and truthfulness. 
Representation Engineering~\citep{olson2024representation} explores editing latent vectors to induce desired outputs, while Rank-One Model Editing (ROME)~\citep{meng2022locating} performs low-rank updates to affect factual memory in LMs. 
AutoSteer~\citep{yang2024autosteer} automates the discovery of steering directions by combining unsupervised clustering and contrastive signals.

Although these approaches demonstrate the effectiveness of activation steering, they typically assume fixed intervention positions and rely on heuristic choices of steering strength. In contrast, our work focuses on addressing these limitations by enabling more adaptive and principled activation steering.
\vspace{-0.1in}
\subsection{Adaptive Control} 
Adaptive control seeks to dynamically adjust the strength or weighting of steering at inference time, rather than relying on fixed heuristics or hand-crafted schedules. In this line of work,
A range of approaches have been proposed in recent work~\citep{ross2022attention,geva2023dissecting,lin2023teaching,li2023steering,wang2025adaptive,sun2024contrastive,subramani2024steering}, which instantiate adaptivity through different signals (e.g., probe predictions, semantic cues, or contrastive objectives) and at different granularities (per layer/head/token).
For example, Adaptive Activation Steering (ACT)~\citep{wang2025adaptive} trains probes to detect steering patterns and adaptively scales intervention strength based on probe outputs. 
SADI~\citep{zhang2025sadi} introduces semantics-adaptive dynamically interventions, adjusting activation vectors according to input semantics, while FairSteer~\citep{nguyen2025fairsteer} detects biased representations during inference and applies debiasing steering vectors only when necessary. 
Contrastive Null Steer~\citep{sun2024contrastive} further improves adaptivity by learning to steer away from undesired directions via contrastive supervision. 
Token-wise Adaptive Modulation (TAM)~\citep{liu2024tokenwise} modulates intervention strength at each token position using auxiliary gating networks, enabling fine-grained control over steering dynamics.


\section{Methodology}
\label{sec:method}
In this section, we present \textsc{PIXEL}, as illustrated in Figure~\ref{fig:overview}: we construct a dual-view, property aligned subspace (Section~\ref{sec:subspace}); derive a closed-form minimal intervention rule augmented with orthogonal residual calibration and lightweight position scanning (Section~\ref{sec:adaptive_alpha}); and establish representation level guarantees for safe, effective steering (Section~\ref{sec:theory}).

\subsection{Dual-View Property-Aligned Subspace}
\label{sec:subspace}

\textbf{Subspace from Property-Aligned Contrastive Pairs.} 
We learn the steering subspace from property-aligned contrastive pairs $(x^+,x^-)$ that differ only in whether the input encourages the target property (e.g., factuality or caution). \emph{Unlike many steering-vector pipelines that construct positives/negatives by concatenating a question with a correct vs.\ incorrect answer, our pairs are built without appending any answers}; the model conditions on input-only variants.

Given a candidate pool $\mathcal{C}$ of property-aligned edits and a small probe set $\mathcal{D}_{\mathrm{probe}}$ with a scalar objective $J(\cdot)$, we evaluate each candidate via the per-example gain
\begin{equation}
g(x,c)=J\!\big(x^{+}(c)\big)-J\!\big(x^{-}(c)\big),
\end{equation}
and retain candidates that consistently yield positive improvements (top-$K$ per $x$ in practice), producing a validated set $\mathcal{C}^{\star}$ selected by empirical improvement rather than manual tuning.

To form a structured steering subspace, we aggregate representation differences across $\mathcal{C}^{\star}$ and $\mathcal{D}_{\mathrm{probe}}$. For an input $x$ of length $p$, let $h_t\in\mathbb{R}^{H}$ denote the residual-stream state at token index $t$ (extracted via residual hooks). We perform two forward passes—on $x^{+}$ and $x^{-}$—and compute the differential signal
\begin{equation}
\Delta h_t(x,c)=h_t\!\big(x^{+}(c)\big)-h_t\!\big(x^{-}(c)\big).
\end{equation}
Stacking $\Delta h_t$ over $t$, $x$, and $c\in\mathcal{C}^{\star}$ and then applying SVD/PCA yields a low-dimensional steering basis.

\noindent\textbf{Dual-View Differential Signal Construction.} 
To ensure robustness and decoding relevance, we propose a dual-view modeling approach for hidden-state differentials induced by property-aligned samples. This includes a \textbf{tail-averaged view}, capturing stable conditioning shifts across multiple prompt tokens, and an \textbf{end-token view}, localized precisely at the prompt boundary, directly influencing early decoding steps.

Formally, given a prompt length $p$, we denote $h_t^{\mathrm{plain}}\in\mathbb{R}^H$ as the hidden state at token $t$ under the plain prompt and $h^{(a)}_t$ as the corresponding state under prompt $a$. The dual-view differentials are defined as:
\[
\Delta^{\mathrm{tail}}(x,a)=\operatorname*{mean}_{t\in\mathrm{Tail}_k(x)}\big(h^{(a)}_{t}-h^{\mathrm{plain}}_{t}\big),\quad
\Delta^{\mathrm{end}}(x,a)=h^{(a)}_{p-1}-h^{\mathrm{plain}}_{p-1},
\]
With an adaptive tail window:
\[
\mathrm{Tail}_k(x)=\{p-k,\ldots,p-1\},\quad k=\mathrm{clip}\!\big(\lfloor 0.1\,p\rceil,\,3,\,8\big),
\]
Ensuring consistent scale across varying prompt lengths. The tail-averaged view reduces noise and alignment artifacts over multiple tokens, reflecting coherent contextual shifts. Conversely, the end-token view precisely captures local changes critical to initial decoding logits and attention distributions.

Practically, both views efficiently reuse hidden states from the same forward passes, imposing negligible overhead. Stacking these dual-view differentials across samples and probe examples forms the data matrix:
\[
X=\big[\;\Delta^{\mathrm{tail}}(x,a)\,;\ \Delta^{\mathrm{end}}(x,a)\;\big]_{(x,a)\in\mathcal{D}_{\mathrm{probe}}\times\mathcal{P}^\star}\in\mathbb{R}^{N\times H}.
\]

Applying weighted PCA of rank $r$ (typically $r=2$) on $X$ yields an orthonormal basis $V=\{\mathbf v^{(1)},\mathbf v^{(2)}\}$. We aggregate these into a unified steering direction:
\[
\mathbf v=\mathbf v^{(1)}+\mathbf v^{(2)},\quad
\mathbf u=\frac{\mathbf v}{\|\mathbf v\|},\quad
c=\|\mathbf v\|.
\]

This stacking-based approach requires no meticulous manual prompt selection, grounding the subspace firmly in empirically validated property-aligned samples.
\subsection{Adaptive Intervention Strength and Semantic Calibration}
\label{sec:adaptive_alpha}

\textbf{Minimal Intervention at a Single Position.} 
To steer representations efficiently toward a target property, we formulate an adaptive steering strength problem as constrained geometric optimization. The objective is to identify the minimal scalar intervention at a given hidden state position, ensuring it meets a predefined cosine similarity threshold relative to the target direction.

Formally, consider hidden state $h_{\ell,t}\in\mathbb{R}^{H}$ at layer $\ell$ and token position $t$, with a target direction $v_{\ell}\in\mathbb{R}^{H}$ of magnitude $c_{\ell}=\|v_{\ell}\|$. Given cosine similarity threshold $s\in(0,1)$, we define the minimal steering strength $\alpha$ by solving:
\[
\min_{\alpha\geq 0}\|\alpha v_{\ell}\|\quad\text{s.t.}\quad\cos(h_{\ell,t}+\alpha v_{\ell},v_{\ell})\geq s.
\]

Let $a_{\ell,t}=\langle h_{\ell,t}, v_{\ell}\rangle$ represent the projection onto the target direction, and $B_{\ell,t}=\sqrt{\|h_{\ell,t}\|^2 - a_{\ell,t}^2}$ denote the orthogonal residual magnitude. This optimization has a closed-form solution:
\[
\alpha_{\ell,t}^{*}(s)=\max\left\{\operatorname{ReLU}\left(-\frac{a_{\ell,t}}{c_{\ell}}\right),\quad \frac{B_{\ell,t}\, s}{c_{\ell}\sqrt{1 - s^{2}}}-\frac{a_{\ell,t}}{c_{\ell}}\right\},
\]
Where the first term ensures non-negative projection calibrations, and the second term guarantees achieving the desired cosine similarity.

This minimal intervention perspective interprets smaller values of $\alpha_{\ell,t}^{*}(s)$ as indicating higher positional sensitivity and efficiency toward alignment.

\noindent\textbf{Orthogonal Residual Calibration.}
The global subspace captures an \emph{attribute axis} (e.g., truthfulness/caution) and broadly aligns semantics, yet sample-level \emph{semantic mismatch} remains and can weaken uniform injection. We therefore construct a \emph{sample-specific residual} orthogonal to the global subspace, combine it with the global direction as a new target, and reuse the minimal-intervention closed form to get the injection strength for all positions in each sample.

\medskip\noindent
\textit{Global subspace and projector.}
At layer $\ell$, let $U_\ell=[v_{\ell}^{(1)},\ldots,v_{\ell}^{(r)}]\in\mathbb{R}^{H\times r}$ denote the global subspace (columns from difference statistics and dimensionality reduction). Define the global probe
\[
\bar v_\ell=\sum_{j=1}^r v_{\ell}^{(j)},\qquad
u_\ell=\frac{\bar v_\ell}{\|\bar v_\ell\|}.
\]
Let $\Pi_{U_\ell}$ denote the orthogonal projector onto $\mathrm{span}(U_\ell)$; in the single-vector case ($r=1$), $\Pi_{U_\ell}=u_\ell u_\ell^{\!\top}$. We write $I-\Pi_{U_\ell}$ for the orthogonal complement.

\medskip\noindent
\textit{Sample-level orthogonal residual}
For a sample $x$, form an \emph{answer-agnostic} positive/negative contrastive pair $(\mathcal{P}^+,\mathcal{P}^-)$ that differs only in attribute polarity while keeping the question context fixed.
Let the sample difference be
\[
\delta_\ell(x)\;=\;h_\ell(\mathcal{P}^+) - h_\ell(\mathcal{P}^-),
\]
where $h_\ell(\cdot)$ denotes the pooled hidden state at layer $\ell$ (e.g., last-token or short tail mean).
Remove the global component and normalize:
\[
\tilde r_\ell(x)=(I-\Pi_{U_\ell})\,\delta_\ell(x),\qquad
\hat r_\ell(x)=\frac{\tilde r_\ell(x)}{\|\tilde r_\ell(x)\|}\ \ (\tilde r_\ell(x)\neq 0),
\]
so that $\hat r_\ell(x)\perp U_\ell$ and serves as a \emph{pure semantic} calibration direction.

\medskip\noindent
\textit{Mixed target direction.}
Combine the global direction with the orthogonal residual using a signed, non-negative weight:
\[
\tilde v_\ell(x;\lambda,\rho)=\bar v_\ell+\lambda\,\rho\,\hat r_\ell(x),
\]
with $\lambda\in\{+1,-1\}$ selecting the calibration orientation to reduce mismatch and $\rho\ge 0$ controlling its magnitude.
If $\tilde r_\ell(x)=0$, we fall back to $\bar v_\ell$.

\medskip\noindent
\textit{Minimal intervention with the mixed target.}
Let $\tilde c_\ell=\|\tilde v_\ell\|$, $\tilde u_\ell=\tilde v_\ell/\tilde c_\ell$, and for position $t$ define
\[
\tilde a_{\ell,t}=\langle h_{\ell,t}, \tilde u_\ell\rangle,\qquad
\tilde B_{\ell,t}=\sqrt{\|h_{\ell,t}\|^2-\tilde a_{\ell,t}^{\,2}}.
\]
Replacing $(v_\ell,c_\ell)$ by $(\tilde v_\ell,\tilde c_\ell)$ in the previous section yields, for $s\in(0,1)$,
\[
\boxed{\;
\alpha_{\ell,t}^{\mathrm{orc}}(s;\lambda,\rho)
=\max\!\left\{
\operatorname{ReLU}\!\Big(-\frac{\tilde a_{\ell,t}}{\tilde c_\ell}\Big),
\ \frac{\tilde B_{\ell,t}\,s}{\tilde c_\ell\sqrt{1-s^{2}}}
-\frac{\tilde a_{\ell,t}}{\tilde c_\ell}
\right\}. \;}
\]

Since $\hat r_\ell(x)\perp u_\ell$, the mixture rotates the target direction within $\mathrm{span}\{u_\ell,\hat r_\ell\}$, preserving the attribute component along $u_\ell$ while calibrating sample-level semantic mismatch via the orthogonal component.


\noindent\textbf{Dynamic Position Scanning.}

\emph{(1) Layer scan at prompt end.}
Let $t_{\mathrm{end}}$ be the last token of prompt, $m(\cdot)$ the metric, $m_0$ the baseline, and
\[
G(\ell)=m \!\big(\text{inject at }(\ell,t_{\mathrm{end}})\big)-m_0.
\]
Let $\ell^\star=\arg\max_\ell G(\ell)$ and $\mathcal{P}=\{\ell:\,G(\ell)>0\}$.
If $\mathcal{P}$ consists of isolated indices, set $\mathcal{S}=\{\ell^\star\}$;
otherwise set $\mathcal{S}$ to the maximal consecutive block in $\mathcal{P}$ containing $\ell^\star$ (simultaneous injection over all $\ell\in\mathcal{S}$).

\emph{(2) Position refinement.}
Fix $\mathcal{S}$ and evaluate $t\in\{t_{\mathrm{end}}-1,\,t_{\mathrm{end}},\,t_{\mathrm{end}}+1\}$; choose
\[
t^\star=\arg\max_{t} m\!\big(\text{inject at all }(\ell\in\mathcal{S},\,t)\big).
\]
We use $(\mathcal{S},t^\star)$ on test set.
\subsection{Representation-Level Guarantees}
\label{sec:theory}

\paragraph{Setup and margin.}
Fix an injection configuration: a nonempty set of layers (sites) $\mathcal{S}\subseteq\{1,\dots,L\}$ and a token position $t^\star$.
For each $\ell\in\mathcal{S}$ and sample $x$, let $w_\ell(x)\in\mathbb{R}^H$ be the unit steering direction used at layer $\ell$.
Denote the pre-injection hidden state at $(\ell,t^\star)$ by $h_{\ell,t^\star}(x)$ and the post-injection state by
\[
h'_{\ell,t^\star}(x)
=
h_{\ell,t^\star}(x)\;+\;\alpha^{*}_{\ell,t^\star}\!\big(h_{\ell,t^\star}(x);\,w_\ell(x),\,s\big)\,w_\ell(x),
\]
where $\alpha^{*}_{\ell,t^\star}(\cdot)$ is the minimal nonnegative strength ensuring
\[
\cos\!\big(h'_{\ell,t^\star}(x),\,w_\ell(x)\big)\ge s,\qquad s\in[0,1).
\]
Write $(z)_+=\max\{z,0\}$ and define the site-wise and averaged (normalized) margins:
\[
\begin{gathered}
\phi_{\mathrm{site}}^{(\ell)}(x)
= \big(\cos(h'_{\ell,t^\star}(x),w_\ell(x)) - s\big)_+ \in [0,1-s],\\[2pt]
\bar\phi_{\mathrm{margin}}(x)
= \frac{1}{|\mathcal{S}|}\sum_{\ell\in\mathcal{S}}\phi_{\mathrm{site}}^{(\ell)}(x),\\[2pt]
\tilde\phi(x)
= \frac{\bar\phi_{\mathrm{margin}}(x)}{1-s} \in [0,1].
\end{gathered}
\]

\begin{theorem}[Generalization of the normalized averaged margin]
\label{thm:gen_margin_fixed_in_method}
Let $\{x_i\}_{i=1}^n$ be i.i.d.\ evaluation samples drawn independently of any data used
to learn the subspace/directions or to choose $(\mathcal{S}, t^\star, s)$.
Then, for this fixed injection configuration, with probability at least $1-\delta$,
\[
\Bigl|\;\mathbb{E}[\tilde\phi(x)]\;-\;\tfrac{1}{n}\sum_{i=1}^n \tilde\phi(x_i)\;\Bigr|
\;\le\;\sqrt{\tfrac{\ln(2/\delta)}{2n}}.
\]
\end{theorem}

\begin{remarkbox}
Because the normalized averaged margin lies between $0$ and $1$, Hoeffding’s inequality yields
distribution-free concentration at the standard square-root-$n$ rate.
The bound depends only on the sample size and the chosen confidence level, and is independent
of the data distribution, hidden dimension, number of sites, and the threshold (after normalization).
\end{remarkbox}

\begin{theorem}[Nonnegativity under per-site minimal intervention]
\label{thm:nonnegative_average_in_method}
Assume $\mathcal{S}\neq\varnothing$ and for each $\ell\in\mathcal{S}$ the strength
$\alpha^{*}_{\ell,t^\star}$ is chosen as the minimal nonnegative value such that
$\cos\!\bigl(h'_{\ell,t^\star}(x),\,w_\ell(x)\bigr)\ge s$.
Then for every sample $x$ and every $\ell\in\mathcal{S}$,
\[
\phi_{\mathrm{site}}^{(\ell)}(x)=\cos\!\bigl(h'_{\ell,t^\star}(x),\,w_\ell(x)\bigr)-s \;\ge\; 0,
\quad\text{and hence}\quad
\bar\phi_{\mathrm{margin}}(x)\;\ge\;0.
\]
\end{theorem}

\begin{remarkbox}
At each selected site we decompose the representation into a component parallel to $w_\ell$
and an orthogonal remainder, then add the smallest nonnegative increment along $w_\ell$
so that the cosine meets the target $s$ (leaving the orthogonal part unchanged).
By monotonicity the site-wise—and hence averaged—margin is nonnegative;
the increment is zero iff the original cosine already meets $s$ and, when needed,
increases strictly with $s$ (see Lemma~\ref{lem:alpha-star} and Lemma~\ref{lem:monotone}).
\end{remarkbox}

\section{Experiments} 

In this section, we present our experimental setup (Section~\ref{sec:exp-settings}), the main results across multiple alignment tasks and models (Section~\ref{sec:exp-main}), ablation studies isolating key components of \textsc{PIXEL} (Section~\ref{sec:exp-ablation}), and a sensitivity analysis of the core steering hyperparameter. (Section~\ref{sec:hyperparam-sensitivity}) 

\begin{table*}[h]
\centering
\setlength{\tabcolsep}{1.5mm}
\caption{Evaluation results on TruthfulQA, BBQ, Refusal, and HelpSteer. The best result is highlighted in bold, and the second-best is underlined. The values reported are the mean performance.}
\resizebox{0.85\textwidth}{!}{%
\begin{tabular}{llllll}
\hline\hline
\textbf{Method} & \multicolumn{2}{c}{\textbf{TruthfulQA}} & \textbf{BBQ} & \textbf{Refusal} & \textbf{HelpSteer} \\
& MC1~($\uparrow$) & MC2~($\uparrow$) & Acc~($\uparrow$) & Sorry~($\uparrow$) & Help.~($\uparrow$) \\
\hline
Llama3-8B-inst. & 28.70$_{\pm 0.21}$ & 45.03$_{\pm 0.28}$ & 0.625$_{\pm 0.011}$ & 0.49$_{\pm 0.02}$ & 3.76$_{\pm 0.03}$ \\
ICL             & 29.60$_{\pm 0.25}$ & 45.61$_{\pm 0.26}$ & 0.638$_{\pm 0.013}$ & \underline{0.52}$_{\pm 0.02}$ & \underline{3.82}$_{\pm 0.02}$ \\
CAA             & 29.64$_{\pm 0.18}$ & 46.95$_{\pm 0.23}$ & 0.646$_{\pm 0.009}$ & 0.49$_{\pm 0.02}$ & 3.77$_{\pm 0.02}$ \\
ITI             & \underline{37.03}$_{\pm 0.20}$ & 53.69$_{\pm 0.29}$ & 0.629$_{\pm 0.012}$ & 0.28$_{\pm 0.02}$ & \underline{3.82}$_{\pm 0.02}$ \\
ReFT            & 30.81$_{\pm 0.16}$ & 48.91$_{\pm 0.22}$ & \underline{0.654}$_{\pm 0.008}$ & 0.45$_{\pm 0.02}$ & 3.78$_{\pm 0.02}$ \\
ACT             & 32.10$_{\pm 0.32}$ & \underline{55.87}$_{\pm 0.40}$ & 0.573$_{\pm 0.017}$ & 0.49$_{\pm 0.04}$ & 3.80$_{\pm 0.04}$ \\
PIXEL(Ours)     & \textbf{46.30}$_{\pm 0.18}$ & \textbf{60.28}$_{\pm 0.22}$ & \textbf{0.699}$_{\pm 0.009}$ & \textbf{0.53}$_{\pm 0.02}$ & \textbf{3.88}$_{\pm 0.01}$ \\
\hline
Qwen2-7B-inst.  & 27.65$_{\pm 0.20}$ & 48.31$_{\pm 0.29}$ & 0.648$_{\pm 0.009}$ & 0.38$_{\pm 0.02}$ & 3.51$_{\pm 0.03}$ \\
ICL             & 28.11$_{\pm 0.24}$ & 49.23$_{\pm 0.27}$ & 0.660$_{\pm 0.010}$ & 0.41$_{\pm 0.02}$ & 3.65$_{\pm 0.03}$ \\
CAA             & 29.71$_{\pm 0.17}$ & 48.15$_{\pm 0.25}$ & \underline{0.663}$_{\pm 0.008}$ & 0.40$_{\pm 0.02}$ & \underline{3.73}$_{\pm 0.02}$ \\
ReFT            & \underline{31.10}$_{\pm 0.15}$ & \underline{49.59}$_{\pm 0.24}$ & 0.650$_{\pm 0.008}$ & \underline{0.42}$_{\pm 0.02}$ & 3.63$_{\pm 0.03}$ \\
ACT             & 33.73$_{\pm 0.34}$ & 50.43$_{\pm 0.38}$ & 0.5625$_{\pm 0.017}$ & 0.39$_{\pm 0.03}$ & 3.60$_{\pm 0.04}$ \\
PIXEL(Ours)     & \textbf{37.33}$_{\pm 0.16}$ & \textbf{55.22}$_{\pm 0.20}$ & \textbf{0.732}$_{\pm 0.006}$ & \textbf{0.66}$_{\pm 0.01}$ & \textbf{3.94}$_{\pm 0.01}$ \\
\hline
Mistral-7B-v0.3 & 22.22$_{\pm 0.28}$ & 36.99$_{\pm 0.33}$ & 0.680$_{\pm 0.014}$ & 0.23$_{\pm 0.03}$ & 3.75$_{\pm 0.03}$ \\
ICL             & 24.31$_{\pm 0.26}$ & 49.68$_{\pm 0.30}$ & 0.688$_{\pm 0.013}$ & 0.24$_{\pm 0.03}$ & \underline{3.77}$_{\pm 0.03}$ \\
CAA             & 32.16$_{\pm 0.19}$ & \textbf{52.50}$_{\pm 0.26}$ & \underline{0.712}$_{\pm 0.009}$ & 0.26$_{\pm 0.02}$ & 3.76$_{\pm 0.02}$ \\
ReFT            & \underline{33.46}$_{\pm 0.17}$ & 50.14$_{\pm 0.28}$ & 0.680$_{\pm 0.011}$ & \underline{0.27}$_{\pm 0.02}$ & \textbf{3.80}$_{\pm 0.02}$ \\
ACT             & 24.15$_{\pm 0.30}$ & 47.21$_{\pm 0.36}$ & 0.5547$_{\pm 0.019}$ & 0.23$_{\pm 0.03}$ & 3.35$_{\pm 0.05}$ \\
PIXEL(Ours)     & \textbf{33.80}$_{\pm 0.12}$ & \underline{50.65}$_{\pm 0.13}$ & \textbf{0.736}$_{\pm 0.008}$ & \textbf{0.54}$_{\pm 0.01}$ & 3.76$_{\pm 0.02}$ \\
\hline\hline
\end{tabular}
}
\label{tab:main-results}
\end{table*}

\subsection{Experimental Settings}
\label{sec:exp-settings}
\noindent{\textbf{Datasets and Metrics.}}
We evaluate PIXEL on four widely used alignment suites, each with its standard metric and judging protocol:
\begin{itemize}
    \item \textbf{TruthfulQA}~\citep{lin-etal-2022-truthfulqa}: multiple-choice factuality. 
    We report \emph{MC1} (top-1 accuracy: whether the true option receives the highest probability) and \emph{MC2} (mean probability mass assigned to the true option across all choices).
    \item \textbf{BBQ}~\citep{parrish-etal-2022-bbq}: stereotypical bias on the disambiguated subset; we report \emph{accuracy}, which reflects answering with the correct, bias-neutral option rather than a stereotype-consistent one.
    \item \textbf{Refusal}~\citep{xie2025sorrybench}: safety refusal on \textit{Sorry-Bench} with automatic judgments by \textsc{Mistral-7B-Instruct-v0.2}; the score reflects the proportion/strength of appropriate refusals to unsafe or malicious requests (higher is better).
    \item \textbf{HelpSteer}~\citep{nguyen2025multi,wang2023helpsteer}: assistant \emph{helpfulness} on a 0–4 scale using \textsc{GPT-3.5-Turbo} as the rater; higher indicates more useful, instruction-following responses.
\end{itemize}

 To verify that steering does not harm utility, we measure accuracy on standard NLP benchmarks:
\begin{itemize}
    \item \textbf{RACE}~\citep{lai-etal-2017-race}: multi-passage reading comprehension from English exams (middle/high school). Each question is a 4-way multiple choice and often requires multi-sentence inference; we report accuracy.
    \item \textbf{MMLU}~\citep{hendrycks2020measuring}: a 57-subject, 4-option multiple-choice suite spanning STEM, humanities, social sciences, and professional topics; it probes world knowledge and reasoning under few-shot prompting. We report average accuracy over subjects.
    \item \textbf{OpenBookQA}~\citep{mihaylov-etal-2018-suit}: 4-way multiple-choice science questions that require combining a provided “open book” of core facts with commonsense/novel inference; we report accuracy.
    \item \textbf{GLUE}~\citep{wang2018glue}: a collection of natural language understanding tasks (e.g., entailment, paraphrase, sentiment) covering sentence- and sentence-pair classification. We follow standard evaluation and report accuracy (task-wise, averaged when aggregated).
\end{itemize}

\noindent{\textbf{Models and Baselines.}}
We consider three representative large language models for experiments:
\begin{itemize}
    \item \textbf{Llama3-8B-Instruct}~\citep{grattafiori2024llama}: an open, instruction-aligned model widely used as an alignment testbed.
    \item \textbf{Qwen2-7B-Instruct}~\citep{team2024qwen2}: a strong general-purpose 7B model with competitive reasoning and safety behaviors.
    \item \textbf{Mistral-7B-v0.3}~\citep{jiang2023mistral7b}: a compact 7B backbone emphasizing efficiency and strong zero-shot performance.
\end{itemize}

We compare PIXEL against prevalent steering paradigms:
\begin{itemize}
    \item \textbf{ICL}~\citep{brown2020language}: prompt-only steering via carefully designed instructions, without parameter or activation edits.
    \item \textbf{ReFT}~\citep{wu2024reft}: representation fine-tuning that learns intermediate activations for attribute alignment.
    \item \textbf{ITI}~\citep{NEURIPS2023_81b83900}: inference-time intervention that perturbs hidden states to reshape the output distribution.
    \item \textbf{CAA}~\citep{rimsky-etal-2024-steering}: contrastive activation addition using differences between positive/negative exemplars.
    \item \textbf{ACT}~\citep{wang2024adaptive}: tuning-free adaptive control that scales activation edits via probe-driven signals at inference.
\end{itemize}

\noindent{\textbf{Experimental Setup.}} All experiments were conducted on NVIDIA RTX 6000 Ada GPUs. For all datasets, subspace extraction used 200 samples. We used subspace similarity thresholds of 0.85 and 0.9. For each configuration, we report the average and standard deviation over three runs with different random seeds.

\begin{table}[h]
\centering

\caption{General capabilities on several benchmarks. }
\begin{tabular}{lcccc}
\hline\hline
\textbf{Method} & \textbf{RACE} & \textbf{MMLU} & \textbf{OpenBookQA} & \textbf{GLUE} \\
\hline
Llama3-8B-instruct & 0.671 & 0.655 & 0.556 & 0.726 \\
ReFT       & \underline{0.677} & \underline{0.651} & \textbf{0.559} & \underline{0.757} \\
ITI        & 0.589 & 0.546 & 0.507 & 0.742 \\
CAA        & 0.671 & 0.648 & \underline{0.557} & 0.738 \\
PIXEL(Ours)       & \textbf{0.690} & \textbf{0.668} & 0.556 & \textbf{0.797} \\
\hline
Qwen2-7B-instruct & 0.625 & 0.695 & 0.606 & 0.825 \\
ReFT       & \textbf{0.644} & \underline{0.698} & \underline{0.613} & 0.770 \\
CAA        & \underline{0.633} & \underline{0.698} & 0.609 & \underline{0.830} \\
PIXEL(Ours)        & 0.625 & \textbf{0.718} & \textbf{0.625} & \textbf{0.871} \\
\hline
Mistral-7B-v0.3 & 0.678 & 0.618 & 0.602 & 0.681 \\
ReFT       & \underline{0.679} & 0.603 & 0.608 & 0.655 \\
CAA        & 0.667 & \underline{0.619} & \underline{0.611} & \textbf{0.693} \\
PIXEL(Ours)       & \textbf{0.697} & \textbf{0.664} & \textbf{0.641} & \underline{0.681} \\
\hline\hline
\end{tabular}
\label{tab:general-capability}
\end{table}

\subsection{Main results}
\label{sec:exp-main}
\noindent{\textbf{Comprehensive gains across heterogeneous evaluation forms.}}
\noindent PIXEL achieves significant improvements under two evaluation paradigms: multiple-choice classification (TruthfulQA MC1/MC2 for factuality; BBQ accuracy for bias) and open-ended generation (auto judges score the rejection of unsafe requests; HelpSteer-Help scores helpfulness on a scale of 0--4). Taking Qwen2-7B as the representative base, PIXEL achieves 37.50/55.22 on MC1 and MC2, 0.732 on BBQ, 0.66 on Rejection, and 3.94 on HelpSteer-Help; these correspond to relative gains of 42\%, 22\%, 15\%, 74\%, and 12\% relative to the base model (26.38/45.41, 0.634, 0.38, 3.51), and improvements of 26\%, 13\%, 15\%, 57\%, and 6\% relative to the strongest competing baseline in this setting (29.83/48.69, 0.636, 0.42, 3.73). We attribute these broad gains to an answer-agnostic approach to constructing guidance: rather than forming pairs by associating known correct and incorrect answers (an approach that entangles the learned signal with format-specific templates and labels), we incorporate generic attribute-related instructions (e.g., \texttt{Avoid repeating myths; respond concisely with verified facts.}) with the original question (e.g., \texttt{What happens if you prick your finger on a spindle?}). This decoupling enables the learned directions to transfer from truthfulness and bias detection in multiple-choice questions to rejection and assistance in open-ended generation. Furthermore, dual-view refinement within a layer (tail-window averaging and end-token differentials) provides complementary information, capturing both stable conditioning shifts and localized changes relevant to initial decoding, thereby supporting comprehensive improvements across evaluation modalities.

\noindent{\textbf{PIXEL maintains model performance on general capability benchmarks.}} To ensure that our steering approach does not compromise the model's inherent language understanding and reasoning capabilities, we evaluate its performance on a range of standard NLP benchmarks, including RACE (reading comprehension), MMLU (multi-task knowledge), OpenBookQA (commonsense reasoning), and GLUE (general language understanding). The results in Table~\ref{tab:general-capability} show that PIXEL not only maintains the performance of the base model but also improves it in some cases, highlighting its precision in attribute alignment without compromising core performance. This stands in stark contrast to baseline methods, which often suffer from performance trade-offs. For example, while CAA relies on explicit positive/negative examples and excels on knowledge-focused closed-set tasks such as MMLU and GLUE (Qwen2 scores 0.830 on GLUE), it exhibits instability on complex reasoning tasks (0.625 on RACE), suggesting that its steering vectors may overfit to simple fact contrasts at the expense of nuanced reasoning paths. ReFT improves one metric (0.677 on RACE for Llama3) but underperforms another (MMLU falls below 0.651 on the original model), indicating limited generalization. Most notably, ITI's fixed-weight intervention leads to a significant drop in knowledge-intensive benchmarks (MMLU drops to 0.546 on Llama3), likely because the coarse-grained activation steering undermines the model's generalizability. In contrast, PIXEL performs position-adaptive steering, applying minimal geometric consistency intervention only at the most acceptable locations within the model's hidden representation. This design ensures minimal disruption to the underlying model's behavior. Intuitively, if alignment with the target attribute can be achieved with a smaller steering strength adjustment, it may interfere less with the model's capabilities on other attributes than achieving the same effect with a larger steering strength.

\subsection{Ablation Studies}
\label{sec:exp-ablation}

\begin{figure*}[h]
  \centering
  \includegraphics[width=\textwidth]{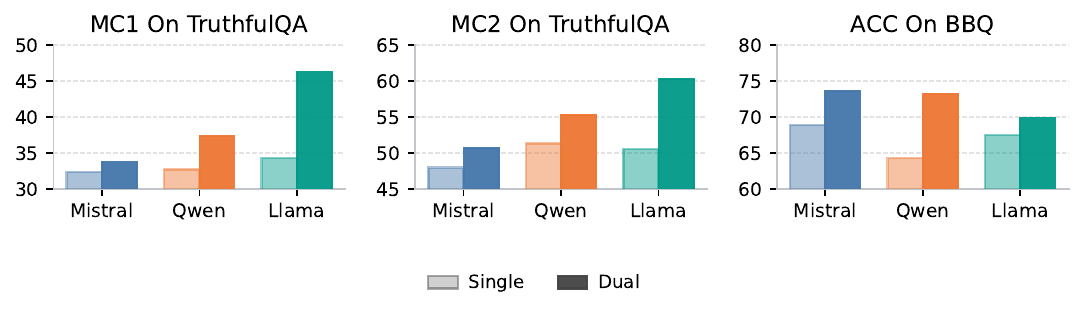}
  \caption{Ablation on Property Subspace Views. (Single vs. Dual) }
  \label{fig:ablation-views}
\end{figure*}

\begin{table}[b]
\centering
\setlength{\tabcolsep}{2.5pt}      
\caption{Comparison of steering strength across datasets. The best result is highlighted in bold.}
\begin{tabular}{lcccccc}
\hline\hline
\textbf{Method} & \multicolumn{2}{c}{\textbf{TruthfulQA}} & \textbf{BBQ} & \textbf{Refusal} & \textbf{HelpSteer} \\
\cline{2-3}
& MC1 ($\uparrow$) & MC2 ($\uparrow$) & Acc ($\uparrow$) & Sorry ($\uparrow$) & Help. ($\uparrow$) \\
\hline
\multicolumn{6}{l}{\textit{Llama3-8B-Instruct}}\\
\hline
w/o adaptive & 35.19 & 51.20 & 0.671  & 0.49 & 3.56 \\
PIXEL        & \textbf{46.30} & \textbf{60.28} & \textbf{0.699} & \textbf{0.53} & \textbf{3.88} \\
\hline
\multicolumn{6}{l}{\textit{Qwen2-7B-Instruct}}\\
\hline
w/o adaptive & 29.48 & 48.65 & 0.673  & 0.42 & 3.59 \\
PIXEL        & \textbf{37.33} & \textbf{55.22} & \textbf{0.732} & \textbf{0.66} & \textbf{3.94} \\
\hline
\multicolumn{6}{l}{\textit{Mistral-7B-v0.3}}\\
\hline
w/o adaptive & 30.09 & 47.47 & 0.7222 & 0.26 & 3.64 \\
PIXEL        & \textbf{33.80} & \textbf{50.65} & \textbf{0.736} & \textbf{0.54} & \textbf{3.76} \\
\hline\hline
\end{tabular}
\label{tab:ablation-adaptive-alpha}
\end{table}

\noindent{\textbf{Validating the Effectiveness of the Proposed Dual-View Property-Aligned Subspace.}} We designed a PIXEL variant (\textbf{Single-View}) that removes the tail average view and uses only the last token difference to construct the attribute subspace. Figure~\ref{fig:ablation-views} compares the performance of Single-View and Dual-View on various datasets. As shown, Dual-View improves TruthfulQA performance across all backbone networks. On MC1, Mistral's score improves from 31.2 to 34.1 (+2.9), Qwen's score improves from 30.5 to 36.4 (+5.9), and Llama's score improves from 32.0 to 46.8 (+14.8). MC2 also showed similar improvements: Mistral's score increased from 52.1 to 54.3 (+2.2), Qwen's score increased from 55.0 to 59.2 (+4.2), and Llama's score increased from 51.3 to 60.5 (+9.2). Dual-view also outperformed Single-View on BBQ (accuracy) across all three backbones. These results suggest that restricting the subspace to terminal label signals underutilizes cues distributed along the tail of the response, thereby limiting attribute alignment for factuality indicators. By aggregating complementary views, the dual-view construction provides a more robust, attribute-aligned global subspace, making the model's activation steering more accurate and stable.

\noindent{\textbf{Evaluating the effectiveness of the adaptive steering strength mechanism.}} To evaluate the impact of steering strength regulation, we designed a PIXEL variant(\textbf{w/o adaptive} ) that disables the adaptive weighting mechanism and instead applies a fixed intervention magnitude $\alpha=1$ across all positions and layers. Table~\ref{tab:ablation-adaptive-alpha} shows a comparison between this fixed-strength variant and the full PIXEL. Across all models and evaluation settings, the adaptive strategy consistently leads to stronger performance. For example, on Llama3-8B-Instruct, using adaptive $\alpha$ yields an MC2 score of 60.28, compared to 51.20 without it; on Qwen2-7B-Instruct, the rejection rate improves from 0.42 to 0.66. These results highlight two core limitations of fixed-strength intervention. First, a uniform intensity can be insufficient at locations where critical steering is required, leading to understeering and limited behavioral change. Second, indiscriminately applying the same magnitude can induce oversteering at places that are semantically inconsistent with the target orientation, leading to poor generalization or unnatural generation.
In contrast, PIXEL utilizes closed-form optimization to compute the minimum required steering magnitude at each location, subject to a cosine similarity threshold relative to the target subspace. This adaptive mechanism enables the model to inject only when necessary and with just enough strength to achieve alignment, balancing effectiveness and safety. The observed improvements demonstrate that precise, geometry-aware interventions are more robust and transferable than coarse, global interventions.

\begin{table}[b]
\centering

\setlength{\tabcolsep}{2.5pt}
\caption{Ablation on sample-level calibration: w/o calibration (global-only subspace) vs. full PIXEL. Best results are in \textbf{bold}.}
\begin{tabular}{lcccccc}
\hline\hline
\textbf{Method} & \multicolumn{2}{c}{\textbf{TruthfulQA}} & \textbf{BBQ} & \textbf{Refusal} & \textbf{HelpSteer} \\
\cline{2-3}
& MC1 ($\uparrow$) & MC2 ($\uparrow$) & Acc ($\uparrow$) & Sorry ($\uparrow$) & Help. ($\uparrow$) \\
\hline
\multicolumn{6}{l}{\textit{Llama3-8B-Instruct}}\\
\hline
w/o calibration & 43.06 & 59.57 & 0.675 & 0.50 & 3.77 \\
PIXEL          & \textbf{46.30} & \textbf{60.28} & \textbf{0.699} & \textbf{0.53} & \textbf{3.88} \\
\hline
\multicolumn{6}{l}{\textit{Qwen2-7B-Instruct}}\\
\hline
w/o calibration & \textbf{37.50} & 55.13 & 0.708 & 0.60 & 3.51 \\
PIXEL          & 37.33 & \textbf{55.22} & \textbf{0.732} & \textbf{0.66} & \textbf{3.94} \\
\hline
\multicolumn{6}{l}{\textit{Mistral-7B-v0.3}}\\
\hline
w/o calibration & 29.17 & 48.46 & 0.7124 & 0.50 & 3.64 \\
PIXEL          & \textbf{33.80} & \textbf{50.65} & \textbf{0.736} & \textbf{0.54} & \textbf{3.76} \\
\hline\hline
\end{tabular}
\label{tab:ablation-calibration}
\end{table}

\noindent{\textbf{Validating the effectiveness of the proposed orthogonal residual calibration mechanism.}} To evaluate the impact of orthogonal residual calibration, we designed a PIXEL variant (\textbf{w/o calibration}) that disables per-sample orthogonal residuals and adjusts only along global, attribute-aligned directions. Table~\ref{tab:ablation-calibration} compares this variant with PIXEL. Disabling calibration leads to consistent performance degradation across models and datasets. For example, on Mistral-7B-v0.3, the MC1 score drops from 33.80 to 29.17 (-13.7\%); on Qwen2-7B-Instruct, the usefulness drops from 3.94 to 3.51 (-11.0\%); and on Llama3-8B-Instruct, the MC1 drops from 46.30 to 43.06 (-7.0\%).
These results demonstrate two limitations of using only the global subspace without calibration. First, it cannot adapt to input-specific semantic shifts, so steering may miss the subspace most relevant to a given cue, suppressing gains in factuality, bias, and safety. Second, it suffers from volatility in open-ended environments: without calibration, interventions may push generations into regions of poor alignment, which manifest as decreased helpfulness or inconsistent rejection. By introducing orthogonal residual calibration, alignment can be improved without changing the injection strength, resulting in more robust and transferable gains.

\subsection{Hyperparameter Sensitivity}
\label{sec:hyperparam-sensitivity}
\begin{figure*}[t]
  \centering
  \includegraphics[width=\textwidth]{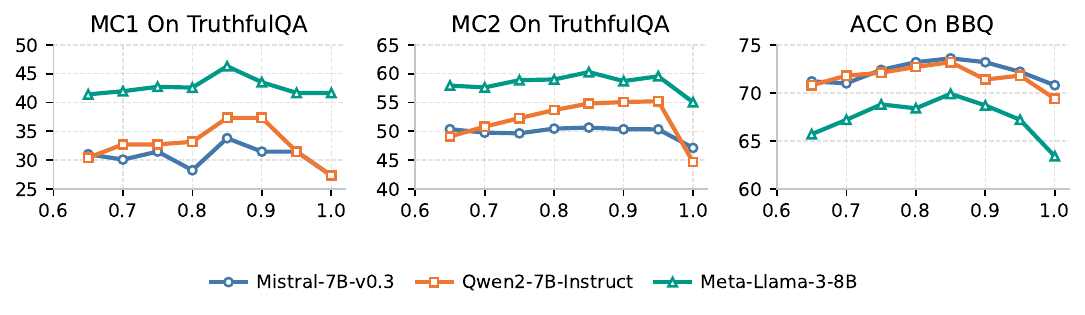}
  \caption{Analysis of $z_{\text{target}}$ sensitivity in PIXEL. }
  \label{fig:threshold-sensitivity}
\end{figure*}
\noindent{\textbf{Threshold Sensitivity in PIXEL}}
We denote the cosine similarity threshold used in the method as $z_{\text{target}}$ during experiments.
We analyze the impact of the cosine similarity threshold $z_{\text{target}} \in [0.6, 1.0]$, which serves as a key hyperparameter in the closed-form optimization for computing minimal intervention strength. This threshold determines how closely the updated activation must align with the target subspace direction, balancing steering effectiveness and representational stability.

To assess its influence, we conduct evaluations on three metrics: TruthfulQA-MC1, TruthfulQA-MC2, and BBQ Accuracy across three backbone models: Llama3-8B-Instruct, Qwen2-7B-Instruct, and Mistral-7B-v0.3. Results are visualized in Figure~\ref{fig:threshold-sensitivity}.We summarize the following observations:
\begin{itemize}
    \item \textbf{Stable behavior across a broad range.} PIXEL achieves strong performance across a wide interval of $z_{\text{target}}$ values, indicating that our method is not overly sensitive to this hyperparameter and maintains robustness without fine-tuning.

    \item \textbf{Under- and over-steering effects.} Extremely low thresholds (e.g., 0.6) tend to permit insufficient alignment, resulting in weak behavioral control. Conversely, overly high thresholds (e.g., 1.0) demand excessive intervention strength, potentially disrupting latent representations and degrading output quality.

    \item \textbf{Model-specific variance.} While Llama3-8B exhibits stable performance across all settings, Mistral-7B shows more pronounced drops at the low end of $z_{\text{target}}$. This suggests that model architecture may influence sensitivity, though PIXEL remains effective across all backbones.
\end{itemize}


\section{Conclusion}

In this paper, we propose Positional Injection via Precision Estimation Levels (PIXEL), a principled, tuning-free framework for positional activation control in large language models, balancing precise control with generalizability. Rather than relying on fixed intervention magnitudes or manually selected injection points, PIXEL learns a robust, attribute-aligned direction by fusing tail-averaged evidence and end-token evidence in a two-view subspace. The minimum required level of control is then determined by applying a closed-form geometric optimization to each layer and each token. To adapt to input-specific semantics without sacrificing global consistency, PIXEL further introduces orthogonal residual calibration, which aligns the global subspace toward a sample-level subspace. Combined with dynamic position scanning, this enables fine-grained, low-overhead interventions. These interventions generalize to heterogeneous objectives (including realism, bias reduction, appropriate rejection, and usefulness) while maintaining performance on standard capability benchmarks. Extensive experiments and ablation studies validate these claims, demonstrating that our geometry-aware, position-adaptive steering approach achieves consistent improvements over strong baselines. In summary, PIXEL provides a scalable foundation for reliable and consistent language generation and a flexible plugin for future multi-attribute control.

\bibliographystyle{ACM-Reference-Format}
\bibliography{iclr2026_conference}

\newpage
\appendix

\section{Mathematical Derivations}
\label{app:proofs-rep}

\subsection{Preliminaries and Lemmas}

\begin{lemma}[Hoeffding’s inequality]
\label{lem:hoeffding}
Let $Y_1,\dots,Y_n$ be i.i.d.\ random variables with $Y_i\in[a,b]$. Then for any $\varepsilon>0$,
\[
\Pr\!\Big(\Big|\tfrac{1}{n}\sum_{i=1}^n Y_i - \mathbb{E}[Y_1]\Big|>\varepsilon\Big)\;\le\;2\exp\!\Big(-\tfrac{2n\varepsilon^2}{(b-a)^2}\Big).
\]
\end{lemma}

\begin{lemma}[Monotonicity of directional cosine under positive shift]
\label{lem:monotone}
Fix a unit vector $w$ and decompose any $h\in\mathbb{R}^H$ as $h=a w + u$ with $a=\langle h,w\rangle$ and $u\perp w$.
For $\alpha\in\mathbb{R}$, define $h'(\alpha)=h+\alpha w$ and $g(\alpha)=\cos(h'(\alpha),w)$.
Then
\[
g'(\alpha)=\frac{\|u\|^2}{\big((a+\alpha)^2+\|u\|^2\big)^{3/2}}\;\ge\;0,
\]
hence $g$ is nondecreasing in $\alpha$.
\end{lemma}

\begin{lemma}[Closed-form minimal strength for $\cos(h+\alpha w,w)\ge s$]
\label{lem:alpha-star}
Let $w$ be unit, $s\in[0,1)$, and write $h=a w+u$ with $u\perp w$.
The minimal nonnegative $\alpha$ such that $\cos(h+\alpha w,w)\ge s$ is
\[
\alpha^{\ast}(h;w,s)\;=\;\Big[\;\frac{s}{\sqrt{1-s^2}}\;\|u\|\;-\;a\;\Big]_+.
\]
\begin{proof}
Let $y=a+\alpha$. Then
\[
\cos(h+\alpha w,w)\;=\;\frac{y}{\sqrt{y^2+\|u\|^2}}.
\]
By Lemma~\ref{lem:monotone}, the smallest $y$ achieving $\cos\ge s$ satisfies equality:
$\,y/\sqrt{y^2+\|u\|^2}=s$.
Solving gives $(1-s^2)y^2=s^2\|u\|^2$, hence $y_{\min}=\tfrac{s}{\sqrt{1-s^2}}\|u\|\ge 0$ (since $s\ge0$).
Therefore $\alpha_{\min}=y_{\min}-a=\tfrac{s}{\sqrt{1-s^2}}\|u\|-a$.
If this is negative, $\alpha^{\ast}=0$ already achieves $\cos(h,w)\ge s$, so taking $[\cdot]_+$ yields the stated formula.

\end{proof}
\end{lemma}

\begin{lemma}[Boundedness of the normalized margin]
\label{lem:bounded}
For any $x$, $\ell$, and $s\in[0,1)$, $\phi_{\mathrm{site}}^{(\ell)}(x)\in[0,1-s]$ and consequently $\tilde\phi(x)\in[0,1]$.
\begin{proof}
Since $\cos(\cdot,\cdot)\in[-1,1]$, we have $\cos(h'_{\ell,t^\star},w_\ell)-s\le 1-s$ and $(\cdot)_+\ge 0$, giving the claimed bounds.
Averaging over $\ell$ preserves $[0,1-s]$, and dividing by $(1-s)$ yields $[0,1]$.
\end{proof}
\end{lemma}

\subsection{Proof of Theorem~\ref{thm:gen_margin_fixed_in_method}}

\begin{proof}
By Lemma~\ref{lem:bounded}, $\tilde\phi(x)\in[0,1]$ for every $x$. Let $Y_i=\tilde\phi(x_i)$.
The assumption that the evaluation samples $\{x_i\}_{i=1}^n$ are i.i.d.\ and independent of any data used to learn directions or choose $(\mathcal{S},t^\star,s)$ ensures $\{Y_i\}$ are i.i.d.\ in $[0,1]$ with mean $\mathbb{E}[\tilde\phi(x)]$.
Applying Lemma~\ref{lem:hoeffding} with $a=0$, $b=1$ gives
\[
\Pr\!\Big(\Big|\tfrac{1}{n}\sum_{i=1}^n \tilde\phi(x_i)-\mathbb{E}[\tilde\phi(x)]\Big|>\varepsilon\Big)\;\le\;2e^{-2n\varepsilon^2}.
\]
Setting the right-hand side to $\delta$ and solving for $\varepsilon$ gives
$\varepsilon=\sqrt{\ln(2/\delta)/(2n)}$, which is the desired bound.
\end{proof}

\subsection{Proof of Theorem~\ref{thm:nonnegative_average_in_method}}

\begin{proof}
For each site $\ell\in\mathcal{S}$, the strength $\alpha^{*}_{\ell,t^\star}$ is the minimal nonnegative value achieving $\cos(h'_{\ell,t^\star}(x),w_\ell(x))\ge s$ by Lemma~\ref{lem:alpha-star}.
Hence
\[
\phi_{\mathrm{site}}^{(\ell)}(x)
=\big(\cos(h'_{\ell,t^\star}(x),w_\ell(x))-s\big)_+
=\cos(h'_{\ell,t^\star}(x),w_\ell(x))-s\;\ge 0.
\]
Averaging over the nonempty $\mathcal{S}$ yields $\bar\phi_{\mathrm{margin}}(x)\ge 0$.
\end{proof}

\section{Implementation Details}
\subsection{Algorithm}

Figure illustrates the overall framework of \textsc{PIXEL}, which combines contrastive subspace learning, position-wise optimization, and semantic calibration for precise, tuning-free intervention. To complement this, we present the full algorithmic pipeline in Algorithm~\ref{alg:pixel}.

The process begins by constructing a \emph{dual-view subspace} using answer-agnostic contrastive pairs. For each pair $(x^+, x^-)$ that differs only in attribute polarity (e.g., factual vs. unfactual, cautious vs. overconfident), we compute residual-state differences from two perspectives: a \textit{tail-averaged view} over the final tokens and an \textit{end-token view} at the prompt boundary. These differences are stacked and projected via principal component analysis(PCA) to yield a low-rank, property-aligned subspace. The top-$r$ directions from this decomposition form the basis $V$, and their sum $\bar{v}$ serves as the global steering probe.

To locate where interventions are most effective, we perform a lightweight scan over candidate layers and prompt-end positions. We select a contiguous block of layers $\mathcal{S}$ that exhibit positive steering gain at the final token $t_{\mathrm{end}}$, and further refine the position $t^*$ via a small windowed search.

At inference time, we perform sample-specific injection using a closed-form optimization. For each sample, we construct a contrastive pair $(\mathcal{P}^+, \mathcal{P}^-)$ and compute the residual difference in pooled activations at each selected layer. We then project this residual onto the orthogonal complement of the global subspace, normalize it, and linearly combine it with the global probe to form a \textit{calibrated direction}.

Given this target vector $\tilde{v}_\ell$, we compute the minimal scalar $\alpha^*_{\ell,t^*}$ required to achieve a specified cosine similarity with the original activation. This is solved via a closed-form geometric rule derived in Section~\ref{sec:adaptive_alpha}. The resulting vector is injected into the hidden state at $(\ell, t^*)$.

This inference-only routine requires no gradient updates and is computationally efficient. By combining property-aware direction learning with closed-form per-site injection, \textsc{PIXEL} achieves controllable attribute alignment with minimal disruption to the model's general behavior.

\begin{algorithm}
\caption{PIXEL: Position-wise Injection with eXact Estimated Levels}
\label{alg:pixel}
\small
\textbf{Require:} Property-aligned contrastive pairs $\mathcal{C}$, probe set $\mathcal{D}_{\text{probe}}$, model $M$, layer set $\mathcal{L}$, prompt end position $t_{\text{end}}$, cosine threshold $s \in (0,1)$, calibration parameters $\lambda \in \{+1,-1\}$, $\rho \ge 0$, PCA rank $r$

\textbf{Ensure:} Steering strengths $\alpha^*_{\ell,t^*}$ and directions $w_\ell(x)$ at selected sites

\begin{algorithmic}[1]
\State \textbf{Dual-View Subspace Construction}
\For{each pair $(x^+, x^-) \in \mathcal{D}_{\text{probe}} \times \mathcal{C}$}
    \State $\Delta^{\text{tail}} \leftarrow \frac{1}{|\text{Tail}_k(x)|} \sum_{t \in \text{Tail}_k(x)} (h_t(x^+) - h_t(x^-))$
    \State $\Delta^{\text{end}} \leftarrow h_{p-1}(x^+) - h_{p-1}(x^-)$
\EndFor
\State Stack all $[\Delta^{\text{tail}}; \Delta^{\text{end}}]$ into matrix $X$
\State $V \leftarrow \text{PCA}(X, r)$ \hfill $\triangleright$ Top $r$ directions
\State $\bar{v} \leftarrow \sum_{j=1}^r v^{(j)}$, $u \leftarrow \bar{v}/\|\bar{v}\|$

\State \textbf{Dynamic Position Scanning}
\For{each layer $\ell \in \mathcal{L}$}
    \State Compute gain: $G(\ell) \leftarrow m(\text{inject at } (\ell, t_{\text{end}})) - m_0$
\EndFor
\State $\ell^* \leftarrow \arg\max_\ell G(\ell)$
\State Select maximal consecutive block $\mathcal{S}$ around $\ell^*$ with $G(\ell) > 0$
\State Evaluate $t \in \{t_{\text{end}}-1, t_{\text{end}}, t_{\text{end}}+1\}$, select optimal $t^*$

\State \textbf{Adaptive Sample-wise Steering}
\For{each test sample $x$}
    \For{each layer $\ell \in \mathcal{S}$}
        \State Build contrastive pair $(\mathcal{P}^+, \mathcal{P}^-)$ for $x$
        \State $\delta_\ell(x) \leftarrow h_\ell(\mathcal{P}^+) - h_\ell(\mathcal{P}^-)$
        \State $\tilde{r}_\ell \leftarrow (I - \Pi_V)\delta_\ell(x)$ \hfill $\triangleright$ Orthogonal residual
        \State $\tilde{v}_\ell \leftarrow \bar{v} + \lambda\rho \cdot \tilde{r}_\ell/\|\tilde{r}_\ell\|$
        \State $c_\ell \leftarrow \|\tilde{v}_\ell\|$, $u_\ell \leftarrow \tilde{v}_\ell/c_\ell$
        \State $a_{\ell,t^*} \leftarrow \langle h_{\ell,t^*}, u_\ell \rangle$
        \State $B_{\ell,t^*} \leftarrow \sqrt{\|h_{\ell,t^*}\|^2 - a_{\ell,t^*}^2}$
        \State $\alpha^*_{\ell,t^*} \leftarrow \max\left\{\text{ReLU}\left(-\frac{a_{\ell,t^*}}{c_\ell}\right), \frac{B_{\ell,t^*} s}{c_\ell\sqrt{1-s^2}} - \frac{a_{\ell,t^*}}{c_\ell}\right\}$
        \State $h_{\ell,t^*} \leftarrow h_{\ell,t^*} + \alpha^*_{\ell,t^*} u_\ell$ \hfill $\triangleright$ Update activation
    \EndFor
\EndFor
\State \textbf{return} Steered model with intervention at $(\mathcal{S}, t^*)$
\end{algorithmic}
\end{algorithm}
\subsection{Datasets and Metrics}
\label{appendix:algorithm-a2}
\noindent We provide detailed descriptions of the datasets and evaluation metrics used in our experiments. 
\paragraph{TruthfulQA} TruthfulQA~\citep{lin-etal-2022-truthfulqa} evaluates a model’s ability to produce truthful and informative responses.  
We report:
\begin{itemize}
    \item \textbf{MC1 (Single-true)}: Accuracy in selecting the single correct answer (highest log-probability among 4–5 candidates).
    \item \textbf{MC2 (Multi-true)}: Normalized probability assigned to all true reference answers.
\end{itemize}

\paragraph{BBQ}The Bias Benchmark for QA (BBQ)~\citep{parrish-etal-2022-bbq} measures social bias in QA outputs across nine social dimensions (e.g., race, gender). We report accuracy: whether the model selects the correct answer.

\paragraph{Sorry-Bench}~\citep{xie2025sorrybench}: Evaluates instruction refusal on harmful inputs using a fine-tuned expert model (Mistral-7B-Instruct-v0.2). We report refusal accuracy based on the model’s ability to reject malicious or unethical instructions.

\paragraph{HelpSteer} ~\citep{wang2023helpsteer} is a human-aligned benchmark for evaluating model helpfulness.We report \textbf{Helpfulness}: Relevance and utility of the response, which is rated by GPT-3.5-Turbo, ranging from 0 (poor) to 4 (excellent), and we report the average for each dimension.

\paragraph{General Benchmarks.}
To verify that steering does not impair general capabilities, we evaluate on standard NLP tasks:
\begin{itemize}
    \item \textbf{RACE}~\citep{lai-etal-2017-race}: reading comprehension; metric: \textit{accuracy}.
    \item \textbf{OpenBookQA}~\citep{mihaylov-etal-2018-suit}: elementary science QA; metric: \textit{accuracy}.  
\end{itemize}

\noindent \textbf{MMLU}~\citep{hendrycks2020measuring} (Massive Multitask Language Understanding) is a challenging benchmark designed to evaluate a model’s world knowledge and problem-solving ability under zero-shot and few-shot settings. It comprises 15,908 multiple-choice questions spanning 57 diverse subjects, including STEM, humanities, social sciences, and professional disciplines such as law and ethics. The tasks vary in difficulty from elementary to advanced levels, making MMLU an ideal benchmark for identifying model weaknesses across both general and specialized domains.

Each subject contains at least 100 test questions, exceeding the length of most human exams. The dataset is split into a few-shot development set (5 questions per subject), a validation set (1,540 questions), and a test set (14,079 questions). The evaluation metric is \textit{average accuracy} across all subjects.

\noindent \textbf{GLUE}~\citep{wang2018glue} (General Language Understanding Evaluation) is a widely used benchmark for evaluating general-purpose language understanding. It consists of nine diverse NLP tasks that span a range of linguistic phenomena, including sentiment analysis, paraphrase detection, textual entailment, and question answering. These tasks collectively assess a model’s ability to perform natural language understanding in varied contexts.

The included tasks are:

\textbf{MNLI} (Multi-Genre Natural Language Inference): Predict entailment, contradiction, or neutrality between premise and hypothesis across multiple domains.

\textbf{QNLI} (Question Natural Language Inference): Convert question answering into an entailment task.

\textbf{QQP} (Quora Question Pairs): Detect if two questions from Quora have the same meaning.

\textbf{SST-2} (Stanford Sentiment Treebank): Classify sentiment in movie reviews as positive or negative.

\textbf{CoLA} (Corpus of Linguistic Acceptability): Judge grammatical acceptability of a sentence.

\textbf{STS-B} (Semantic Textual Similarity Benchmark): Score sentence pairs on semantic similarity.

\textbf{MRPC} (Microsoft Research Paraphrase Corpus): Determine if two sentences are paraphrases.

\textbf{RTE} (Recognizing Textual Entailment): Binary entailment classification from multiple datasets.

\textbf{WNLI} (Winograd NLI): Resolve coreference in complex pronoun cases.

\end{document}